\def\year{2020}\relax
\documentclass[letterpaper]{article}
\usepackage{aaai20}  
\usepackage{times}  
\usepackage{helvet} 
\usepackage{courier}  
\usepackage[hyphens]{url}  
\usepackage{graphicx}
\usepackage{graphics}
\usepackage{multirow}
\usepackage{algorithm}
\usepackage{algpseudocode}
\usepackage{float}
\usepackage{numprint}
\usepackage{ctable}
\usepackage{graphicx}  
\usepackage{amssymb,amsmath,amsfonts,bbm,subfigure}
\newcommand{\eg}{e.g.}

\newcommand{\etal}{et~al.}
\DeclareMathOperator{\E}{\mathbb{E}}

\title{DefogGAN: Predicting Hidden Information in the StarCraft Fog of War\\with Generative Adversarial Nets}
\author{Yonghyun Jeong, Hyunjin Choi, Byoungjip Kim, Youngjune Gwon\\
Samsung SDS\\
}

\begin{document}

\maketitle

\begin{abstract}
We propose DefogGAN, a generative approach to the problem of inferring state information hidden in the fog of war for real-time strategy (RTS) games. Given a partially observed state, DefogGAN generates defogged images of a game as predictive information. Such information can lead to create a strategic agent for the game. DefogGAN is a conditional GAN variant featuring pyramidal reconstruction loss to optimize on multiple feature resolution scales. We have validated DefogGAN empirically using a large dataset of professional StarCraft replays. Our results indicate that DefogGAN can predict the enemy buildings and combat units as accurately as professional players do and achieves a superior performance among state-of-the-art defoggers. 
\end{abstract}

\section{Introduction}
The success of AlphaGo~\cite{silver1} has brought a significant attention for artificial intelligence in games (game AI). Agents trained by deep reinforcement learning have demonstrated hands-down victories over expert human players in classic games such as Chess~\cite{silver2}, Go~\cite{silver1}, and Atari~\cite{mnih}. With more complex setting, real-time strategy (RTS) games serve a means to evaluate state-of-the-art learning algorithms. Game AI today opens up new opportunities and challenges for machine learning. The benefits of developing game AI are widespread beyond gaming applications. The exploration to adopt an intelligent agent in science (\eg, predicting protein folding in organic chemistry~\cite{alphafold}) and enterprise business service (\eg, chatbots~\cite{chatbot}) is making to enter a new era for game AI. 

In this paper, we describe DefogGAN that takes a generative approach to compensate imperfect information presented to a gamer due to the fog of war. We use StarCraft, an RTS game featuring three well-balanced races for a gamer to choose and build substantially different playing styles and strategies. StarCraft remains a popular E-sport after more than two decades of the original release. In a daunting aim for our game AI to conquer highly-skilled human players, we train our DefogGAN with more than 30,000 episodes of expert and professional human replays. Such aim has been notoriously difficult for StarCraft whose long withstanding popularity has compounded a broad range of adept game tactics in addition to micro-control techniques~\cite{ontanon} widespread in the E-sport scenes and \texttt{Battle.net}.

The fog of war refers to the lack of vision and information on an area without a friendly unit around it, including all regions that have been previously explored but left unattended currently. Partially Observable Markov Decision Process (POMDP)~\cite{pomdp} best describes the fog of war problem. In general, POMDP gives a practical formulation for most real-world problems characterized by having many unobserved variables. For game AI, solving a partial observation problem is essential to improving its performance. In fact, many existing approaches to design intelligent game AI often suffer from the partial observation problem~\cite{macro}. Recently, generative models are used to alleviate the uncertainty of partial observations. The agent's performance is enhanced from taking advantage of the (predictive) results obtained through a generative model~\cite{synnaeve1,kahng}. The generative approach, however, cannot fully match highly skillful scouting techniques of a top-notch professional human player. 
\begin{figure}[t]
\centering
\includegraphics[width=0.40\textwidth]{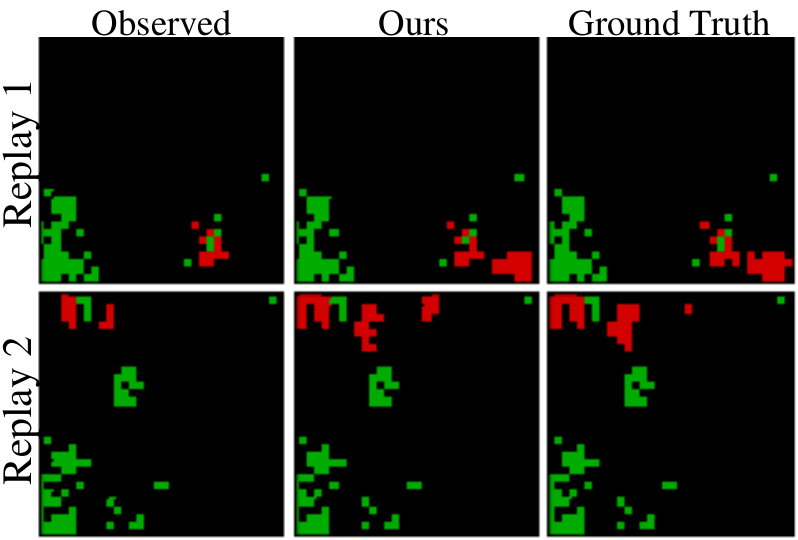}
\caption{Comparison of DefogGAN prediction to ground truth. Friendly and enemy units are represented as green and red in the map (black). The unobserved enemy units are predicted by DefogGAN.}
\label{fig:intro_fig} 
\end{figure}

StarCraft provides a great platform to study complex POMDP problems related to game AI. We set up DefogGAN to accurately predict the state of an opponent hidden in the fog using the \emph{realistic} information generated, thanks to generative adversarial nets (GANs)~\cite{gan}. We find empirically that GANs generate more realistic images than variational autoencoders (VAEs)~\cite{vae}. To generate a defogged game state, we have modified the original GAN generator into an encoder-decoder network. 

In principle, DefogGAN is a variant of conditional GAN~\cite{cgan}. Utilizing skip connections, the DefogGAN generator is trained on residual learning from the encoder-decoder structure. In addition to the GAN adversarial loss, we set up a reconstruction loss between fogged and defogged game states to emphasize the regression of unit positions and quantities. This paper makes the following contributions.
\begin{itemize}\itemsep0em
\item We develop DefogGAN to resolve a fogged game state into useful information for winning. DefogGAN makes one of the earliest GAN-based approaches to cope with the StarCraft fog of war;
\item Using skip connections for residual learning, we have set up DefogGAN to contain past  information (sequence) in a feedforward manner without introducing any recurrent structure, making it suitable for real-time uses;
\item We empirically validate DefogGAN in ablation study and other settings such as testing against extracted game intervals and the current state-of-the-art defog strategy.
\end{itemize}
Our dataset, source code, and pretrained networks are available online for public access.\footnote{\url{https://github.com/TeamSAIDA/DefogGAN}}
\section{Related Work}

\subsection{StarCraft AI}
StarCraft is an immensely successful RTS game developed by Blizzard Entertainment. Since its original release in 1998, StarCraft has attracted professional E-sport leagues and millions of amateur enthusiasts worldwide. Consisting of three fictional races, namely Terran, Protoss, and Zerg, StarCraft is considered as one of the most well-balanced online games ever created. The combinatorial complexity of player actions is extremely high, although at a high level, winning conditions for StarCraft can be built upon the military power and an economy accumulated by the player.

StarCraft AI has a long history, reflecting a number of different playing styles. Ontanon \etal~\shortcite{ontanon} point out that StarCraft playing essentially comprises two tasks. First, micro-management refers to the ability to control units individually. Good micro-management can keep a player's worker and combat units alive for a long time. Secondly, macro-management is the ability to produce units and expand the production facilities into regions other than the start location. 

Defogging can be crucial to both micro- and macro-aspects of the game. Better estimation of hidden areas in the map will help win combats while the player has a higher chance of making the right decision for the future. A poor observation in general can hurt macro-management~\cite{particle,macro}. Scouting is the most straightforward defogging technique~\cite{park2012prediction,si2014scouting}. Interestingly, Justesen \& Risi \shortcite{justesen2017learning} propose a deep learning-based approach to learn the opponent status from units and upgrades information. Generative models give a new class of prediction techniques in StarCraft AI. The convolutional encoder-decoder (CED) model~\cite{synnaeve1,kahng} can be used to recover information hidden in the fog. Synnaeve~\etal~\shortcite{synnaeve1} find beneficial to use a convolutional encoder and a convolutional-LSTM encoder. Our approach of using GAN to generate hidden information as a predictive measure is new to the literature.
\subsection{Generative Adversarial Nets (GAN)}
Goodfellow~\etal~\cite{gan} introduce GAN to generate data from probabilistic sampling. GAN constitutes two neural nets, a generator $G$ and a discriminator $D$, trained in the competition described by a minimax game:
\begin{align}\nonumber
\underset{G}{\min}\,\underset{D}{\max} \E_{x \sim p_{\textrm{real}}}[\log(D(x))] + \E_{z \sim p(z)}[\log (1 - D(G(z)))]
\end{align}

Radford, Metz, and Chintala \shortcite{dcgan} have proposed DCGAN that uses a deep convolutional neural net as $G$. Vanilla GAN is trained on the Jensen-Shannon divergence (JSD), which can cause the vanishing gradient and mode collapse problems. WGAN~\cite{wgan} proposes the use of the Wasserstein-1 metric to improve the vanilla GAN problems. Gulrajani~\etal~\shortcite{wgangp} propose WGAN-GP having a gradient penalty that has a similar effect as the weight clipping. Zhao~\etal~\shortcite{ebgan} introduce Energy-based GAN (EBGAN) using an autoencoder. Berthelot, Schumm, and Metz \shortcite{began} propose BEGAN that combines the WGAN and EBGAN ideas. We will experimentally compare the GAN variants for defogging performances.

\subsection{Generative Approaches for Defogging}
The fog of war problem is similar to inpainting~\cite{nazeri2019edgeconnect} and denoising~\cite{vae}. However, there are three key differences. First, the enemy units may be hidden even in the presence of the friendly units, so defogging must predict the location and the number of each enemy unit type in a 2D grid space up to 4096 $\times$ 4096. Secondly, defogging is a regression problem, which must infer the number of units in the entire area based on a partial observation. Lastly, the problem is not just to generate an image based on the masked (fogged) image. Defogging must indicate the grid where a unit of interest is likely to exist.

\begin{figure*}[h]
    \includegraphics[width=\textwidth]{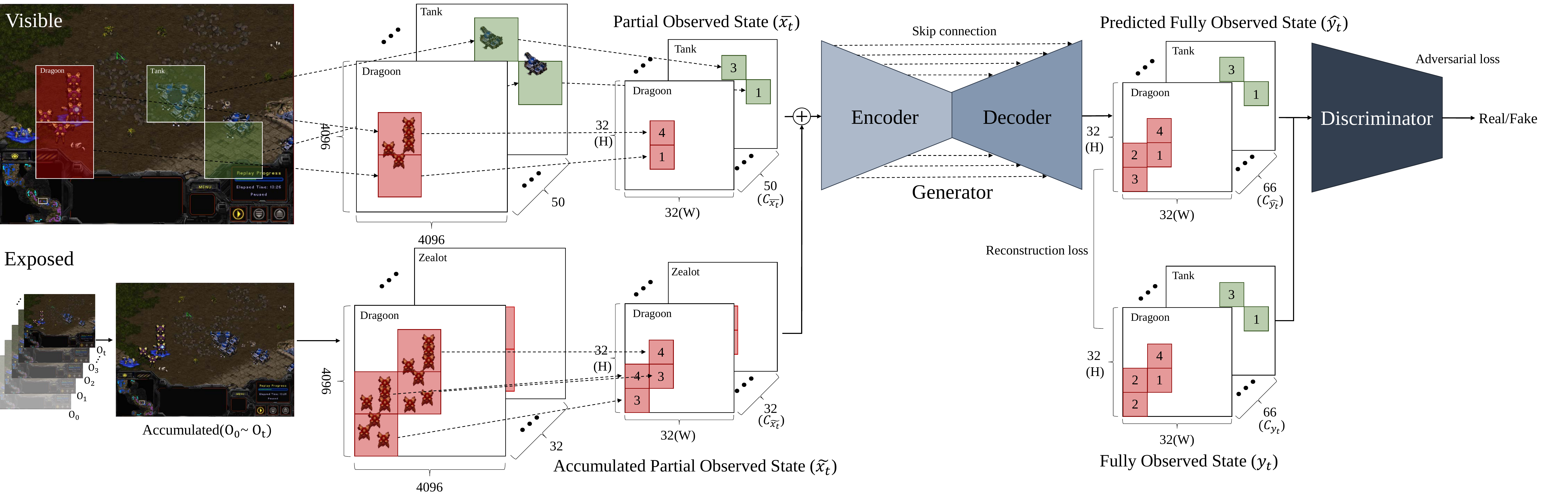}
    \caption{Architectural overview of DefogGAN.}
    \label{fig:defoggan_arch}
\end{figure*}

\section{DefogGAN}
This section presents DefogGAN, explaining its architecture and objective functions. We also describe our implementation details. 

\subsection{Overview}
DefogGAN generates a fully observed (defogged) state from a partially observed (fogged) state at a time $t$. For StarCraft, a fully observed state includes the exact locations of all friendly and enemy units at a given time. Figure~\ref{fig:defoggan_arch} presents DefogGAN. Feature maps computed on the current partially observed state input are sum-pooled. Feature maps on the past observations are accumulated and concatenated to the current before entering the generator. The reconstruction loss between the predicted and the actual fully observed states and the discriminator adversarial loss are used to train the generator. 

% ============
% Notations
% ============
\subsection{Notation}
We denote $y_t$ a ground-truth fully observed state at time $t$, consisting of the exact locations of all units in the game. It is represented as a three-dimensional array of width, height, and channels. Each unit type makes up a channel, and the size of a raw game image in StarCraft is $4096\times4096$ pixels. With 66 unit types, a 1-vs-1 StarCraft game state is $4096\times4096\times66$. We use $\hat{y}_t$ for a predicted fully observed state. A partially observed state at a time $t$ is $\bar{x}_t$. In StarCraft, friendly units are always visible, making the half of the channels in the input fully observed. Ignoring the enemy buildings, which are static units, a partially observed state is an array of size $4096\times4096\times50$. Here, 50 channels include 34 channels for friendly units and 16 channels for enemy combat units. Partially observed states accumulated until time $t$ is denoted by $\tilde{x}_t$. Accumulated partial observations, however, include enemy buildings and exclude friendly units, which are already a part in the current partial observation. This results in an array of size $4096\times4096\times32$ for $\tilde{x}_t$. Combining $\bar{x}_t$ and $\tilde{x}_t$, a concatenated total input $x_{t}$ is applied to DefogGAN.

% ===================================
% (1) Accumulated Partial Observation
% ===================================
\subsection{Accumulating Partial Observations}
Unlike vanilla GAN that generates an image $x_{\textrm{fake}}$ from a latent variable, Defog needs to generate a defogged observation $\hat{y}_{t}$ given a partial observation $\bar{x}_{t}$. Defog has an autoencoder generator instead of a deconvolutional net.
\[
    f(\bar{x}_{t}) = \hat{y}_{t} = G(\bar{x}_{t}) = \textrm{Dec}(\textrm{Enc}(\bar{x}_{t}))
\]

If a partially observed state $\bar{x}_{t}$ lacks temporal information about moving units, it would be insufficient to learn how to generate a fully observed state $y_{t}$. Accumulated partial observation $\tilde{x}_{t}$ facilitates such temporal information. Later in the paper, we show that using accumulated partial observation as an input increases precision and recall. DefogGAN takes in concatenated $\bar{x}_{t}$ and $\tilde{x}_{t}$:     
\[
    x_{t} = \bar{x}_{t} \oplus \tilde{x}_{t}.
\]

Note that we use downsampled $x_{t}$ and $y_{t}$. Since the size of a raw state is too large, we reduce it to 32 x 32. More specifically, as shown in Figure~\ref{fig:defoggan_arch}, a partially observed state $\bar{x}_{t}$ is now an array of size ($32 \times 32 \times 66$). The downsampling allows DefogGAN to efficiently learn how to generate a fully observed state while preserving semantic information of a state~\cite{synnaeve1,kahng}.    

For using temporal information, we could use a recurrent neural net. Using a recurrent neural net, however, comes with some disadvantages such as information dilution and gradient vanishing. Since StarCraft has a relatively long playing time, recurrent nets in general should take too many frames (\eg, to infer game states for a 10-minute duration, 14,400 frames are needed). Our DefogGAN approach has opted for stacking past partial observations onto the current~\cite{mnih}. By incorporating accumulated partial observation $\bar{x}_{t}$, we derive the adversarial objectives
\begin{align}
    \mathcal{L}_{G}^{'} &= \mathcal{L}_{adv}\nonumber\\
   & = \E_{\bar{x}_{t} \sim X_{par}, \tilde{x}_{t} \sim X_{acc}}[\log(1 - D(G(\bar{x}_{t} \oplus \tilde{x}_{t})))]
\end{align}

\begin{align}
    \mathcal{L_{D}} = &-\E_{y \sim Y}[\log(D(y_{t}))]\nonumber\\
    &-\E_{\bar{x}_{t} \sim X_{par}, \tilde{x}_{t} \sim X_{acc}}[\log(1 - D(G(\bar{x}_{t} \oplus \tilde{x}_{t})))]
\end{align}

% =================================
% (2) Pyramidal Reconstruction loss
% =================================
\subsection{Pyramidal Reconstruction Loss}
\begin{figure}[h!]
\centering
\includegraphics[width=0.48\textwidth]{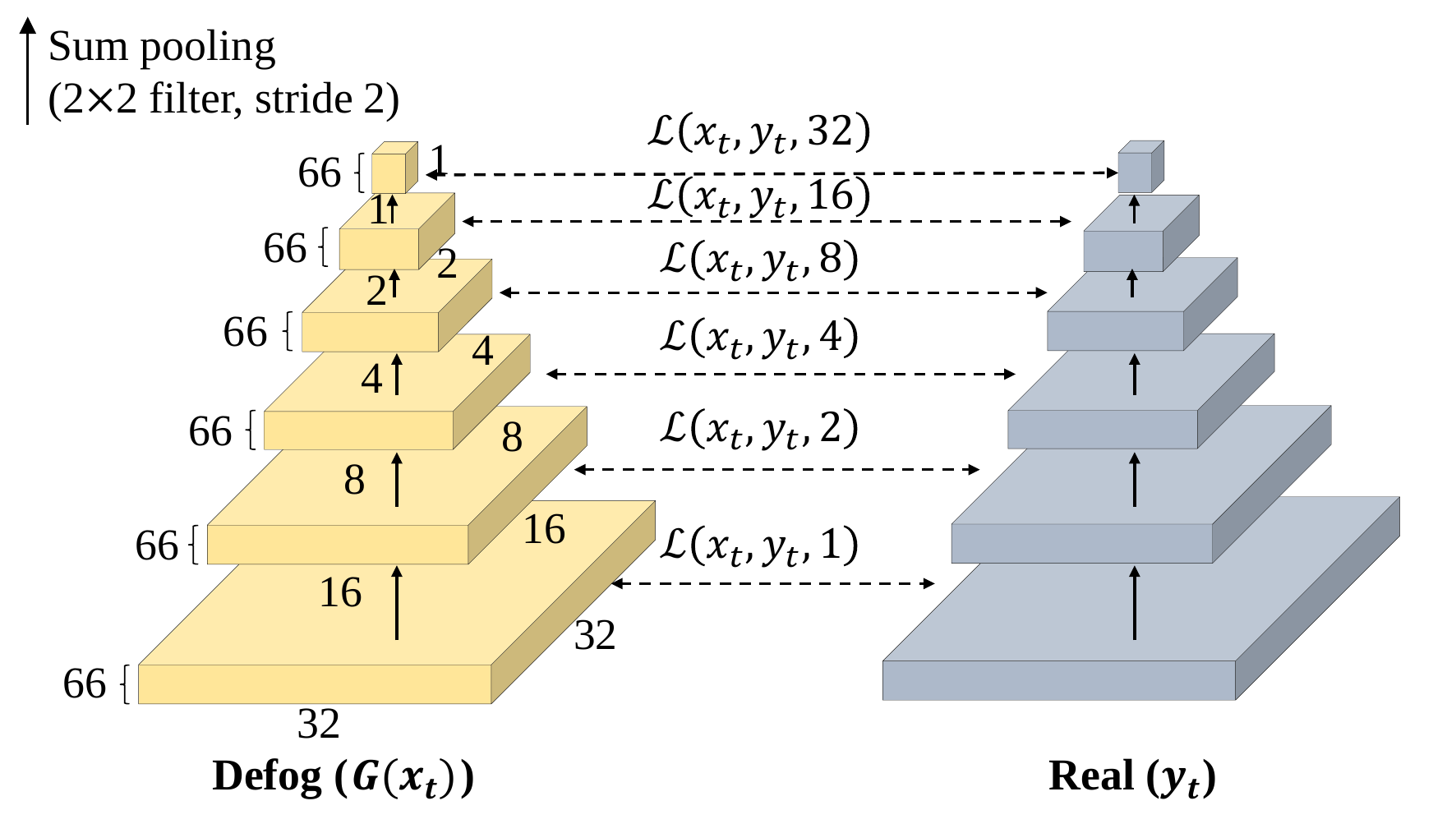}
\caption{Pyramidal reconstruction loss. The pyramid on the left shows the generation of $\hat{y}_{t}$ and the reduction of resolution through sum pooing. The pyramid on the right represents sum pooing of the ground truth $y_{t}$. The mean squared error (MSE) can be measured between the same resolutions. Note that pyramidal loss function $\mathcal{L}(x_{t},y_{t},s)$ measures the MSE with a stride $s$.}
\label{fig:pyramid} 
\end{figure}

We train the generator $G(\bar{x}_{t} \oplus \tilde{x}_{t})$ by minimizing the reconstruction loss between a generated state $\hat{y}_{t}$ and the ground truth $y_{t}$. To further enhance the generator, we introduce pyramidal reconstruction loss as a sum of the MSE between multiple levels of pooling having different sizes ($H$, $W$). Figure~\ref{fig:pyramid} illustrates pyramidal reconstruction loss.
% =====================================
% Figure: Pyramidal Reconstruction Loss
% =====================================

Multiple predictions at different scale are generated by sum pooling. By adjusting filter and stride sizes, sum pooling can generate multiple predictions in a pyramidal shape. More specifically, for a given feature map $m_{t}$, the sum pooling function $\textrm{sumpool}(m_{t}, s)$ creates $m^{'}_{t}$ with a stride $s$ and a filter size $s$. This can be formulated as follows
\[
    \textrm{sumpool}(m,s) = m^{'}_{t}(i,j) = \displaystyle\sum_{h=s\cdot j}^{s\cdot(j+1)-1}\displaystyle\sum_{w=s\cdot i}^{s\cdot (i+1)-1}m_{t}(w,h),
\]
where $(w, h)$ is the coordinates of $m_{t}$, and $(i, j)$ is the coordinates of $m^{'}_{t}$.

At each generation, $\mathcal{L}_{dist}$ is evaluated as follows.
\[
    \mathcal{L}_{dist}(x_{t},y_{t},s) = \E\left[\left \|\textrm{sumpool}(y_{t},s) - \textrm{sumpool}(G(x_{t}),s)\right\|_2^2\right]
\]

For the resolution $r$ (using $r = 32$) to be reduced, pyramidal reconstruction loss is evaluated by 
\begin{align}
    \mathcal{L}_{rec} = \sum_{i=0}^{\log_{2}r}w_i\cdot \mathcal{L}_{dist}(x_{t}, y_{t}, 2^i).
\end{align}

A scaling factor $w_{i}$ adjusts the loss values at different scales. It is defined
\[
    w_{i} = \displaystyle \frac{4^{-i}}{\sum_{k=0}^{\log_{2}r}4^{-k}}.
\]

The proposed pyramidal reconstruction loss allows DefogGAN to learn the total number of units with the lowest resolution of $1 \times 1$. Finally, by incorporating the reconstruction loss, the generator loss of DefogGAN is extended as
\[
    \mathcal{L}_{G} =  \lambda_{adv} \mathcal{L}_{adv} + \lambda_{rec} \mathcal{L}_{rec}.
\]
% ======================================
% (3) U-Net-based Conditional Generation
% ======================================
\subsection{Observation Preserving Connection}
The DefogGAN encoder and decoder are connected in a symmetrical structure. We add residuals between the encoder and decoder at each layer to maintain the parts that have been seen already. By doing so, the generator learns the parts that are hidden in the fog~\cite{resnet,isola2017image}. Through the encoder network, the compressed feature is well-communicated to the decoder for efficient learning. In particular, the observation preserving connections that tie the beginning and the end convey the information that has been observed already. This allows DefogGAN to focus on the information of the units that needs to be inferred. That is, the generator $F(x_{t})$ learns by observing informational connection with $G(x_{t})$ less the observed informational connection $x_t$:
\[
    F(x_{t}) = G(x_{t}) - x_{t}.
\]

%================
% Total Objective
%================
\subsection{Total Objective}
The total objective of DefogGAN is
\begin{align}
    \mathcal{L}_{D} = &-\E_{y \sim Y}[\log(D(y_{t}))]\nonumber\\
        &-\E_{\bar{x}_{t} \sim X_{par}, \tilde{x}_{t} \sim X_{acc}}[\log(1 - D(G(\bar{x}_{t} \oplus \tilde{x}_{t})))],\nonumber\\
    \mathcal{L}_{G} = &~\lambda_{adv} \mathcal{L}_{adv} +  \lambda_{rec} \mathcal{L}_{rec}\nonumber.
\end{align}

Note hyperparameters $\lambda_{adv}$ and $\lambda_{rec}$. In this paper, we use $ \lambda_{rec} = 0.999$ and $\lambda_{adv} = 0.001$.

% =========
% Training
% =========
\subsection{Training}
The overall training procedure of DefogGAN is presented in Algorithm~\ref{alg:alg}. We use Adam~\cite{adam} for training both discriminator and generator.

% ==========
% Algorithm
% ==========
\begin{algorithm}[h!]
\caption{Training the DefogGAN model}
\label{alg:alg}
\begin{algorithmic}

\State $\theta_{G},\theta_{D} \gets $ initialize network parameters
\State $\lambda_{rec}=0.999,\lambda_{adv}=0.001, r=32,epoch =0$
\Repeat
      \State $X\gets $ batch from dataset
      \State $\hat{Y}\gets G(X)$
      \State $Y^{'}\gets Y,\hat{Y}$
      \State $p_{data} \gets D(Y^{'})$
      \State $p_{g} \gets D(\hat{Y})$
      \State $\mathcal{L}_{rec}\gets Eq.(3)$
      \State $ \mathcal{L}_{D} \gets Eq.(2)$
       \State $\mathcal{L}_{adv} \gets Eq.(1)$
      \State $\mathcal{L}_{G} \gets \lambda_{rec}\mathcal{L}_{rec} + \lambda_{adv}\mathcal{L}_{adv}$
      \State //Update parameters according to gradients
      \State $\theta_{G}  \gets -\nabla_{\theta_{G}}\mathcal{L}_{G}$
      \State $\theta_{D}  \gets -\nabla_{\theta_{D}}\mathcal{L}_{D}$
      \State $epoch \gets epoch + 1$
\Until $epoch = 1000$
\end{algorithmic}
\end{algorithm}

% =================
% Implementation
% =================
\subsection{Implementation}

\subsubsection{Generator}
The DefogGAN generator follows the style of the VGG network~\cite{simonyan2014very}. The filter size is fixed at $3 \times 3 $. The number of filters doubles when the feature map size is reduced by half. DefogGAN does not use any spatial pooling or fully-connected layers but uses convolutional layers to preserve spatial information from input to output.

The DefogGAN generator consists of encoder, decoder, and a channel combination layer. The encoder uses $32 \times 32 \times 82$ input and extracts semantic features hidden in the fog by convolutional neural networks (CNNs). Each convolutional layer uses batch normalization and rectified linear unit (ReLU) to make the nonlinear conversion possible~\cite{ioffe2015batch,nair2010rectified}.

The decoder generates predictive data using semantically extracted encoder features. The decoding process reconstructs data into a high dimension, and the inference is done using the transposed convolution operation. The decoder produces the same output shape as the input shape. We do not use as many convolutional layers as ResNet, considering the speed of learning due to the large initial channel size~\cite{resnet}.

The final channel combination layer consists of a single convolutional layer, which combines the 82 channels of accumulated partial observations $C_{\tilde{x}_{t}}$ and $C_{\bar{x}_{t}}$ to obtain 66 channels $C_{\hat{y}_{t}}$ of information to predict. This infers $\hat{y}_{t}$.

\subsubsection{Discriminator}
The DefogGAN discriminator is similar to that of DCGAN~\cite{dcgan}. Three convolutional layers are used with a leaky ReLU activation function. Dropout is used for all layers instead of batch normalization. Through the fully connected final layer, predicting real or fake can be learned.
\section{Experiments}
% ========
% Dataset
% ========
\subsection{Dataset}
We have collected a large dataset of more than 33,000 replays of professional StarCraft players. Our experiments utilize replay log files, which contain detailed unit information. For each frame, a concatenated partial observation ($x_{t}$) of the fogged map $H \times W \times  (C_{\bar{x}_{t}} \oplus C_{\tilde{x}_{t}})$ exists along the corresponding ground truth ($y_t$) in $H \times W \times C_{y_t}$. From each episode, we have decided to only use a portion from the 7th to the 17th minutes. This is because high-level units in a StarCraft game start to appear in about 7 minutes. Also, the game typically finishes in 10 to 20 minutes~\cite{stardata} although not many replays are of more than 17 minutes of duration. Our dataset comprises 496,830 frames. We use 80\% of the data for training, 10\% validation, and 10\% testing. 

Table \ref{tab:confusion_1} summarizes the DefogGAN input-output statistics, including partially observed states $\bar{x}_{t}$, accumulated partially observed states $\tilde{x}_{t}$, and ground truth $y_{t}$. On average, 54\% of the total number of units are seen in partial observation, and 83\% are seen in accumulated partial observation. Note that accumulated partial observation causes a type 1 error (i.e., false positive) because accumulated states contain the previous locations of moving units that are obsolete at the current time. Given this output space, the defog problem is to select an average of 141 spaces out of 67,584 ($32 \times 32 \times 66$) spaces possible.

% ===============================
% Table: Dataset Confusion Matrix
% ===============================
\begin{table}[h!]
\centering
\small
\caption{Confusion matrix of $\bar{x}_{t}$ and $\tilde{x}_{t}$. Using test data (more than 10,000 frames), average numbers are shown.}
\label{tab:confusion_1}
\begin{tabular}{cclrr}
\specialrule{.1em}{.05em}{.05em}
                                                                                                    &           & \multicolumn{2}{c}{$y_{t}$ (GT)}      &\multicolumn{1}{c}{\multirow{2}{*}{Total}}\\ 
\cline{3-4}
                                                                                                    &           & \multicolumn{1}{c}{Exist}    & Not exist  &\multicolumn{1}{c}{}\\ 
\hline
\multicolumn{1}{c}{\multirow{2}{*}{\begin{tabular}[c]{@{}c@{}} $\bar{x}_{t}$ (partial)\end{tabular}}} & Exist     & \multicolumn{1}{r}{81.58} & 0         & 81.58\\ 
\cline{2-5} 
\multicolumn{1}{c}{}                                                                               & Not exist & \multicolumn{1}{r}{59.35} & 67443.07 & 67502.42\\ 
\hline
\multicolumn{2}{c}{Total}& \multicolumn{1}{r}{140.93} & \multicolumn{1}{r}{67443.07} & 67584.00 \\
\specialrule{.1em}{.05em}{.05em}
\specialrule{.1em}{.05em}{.05em}
\end{tabular}
\begin{tabular}{cclrr}
                                                                                                    &           & \multicolumn{2}{c}{$y_{t}$ (GT)}        &\multicolumn{1}{c}{\multirow{2}{*}{Total}}\\ 
\cline{3-4}
                                                                                                    &           & \multicolumn{1}{c}{Exist}    & Not exist  &\multicolumn{1}{c}{}\\ 
\hline
\multicolumn{1}{c}{\multirow{2}{*}{\begin{tabular}[c]{@{}c@{}} $\tilde{x}_{t}$ (accum.)\end{tabular}}} & Exist     & \multicolumn{1}{r}{109.49} & 7.30         & 116.79\\ 
\cline{2-5} 
\multicolumn{1}{c}{}                                                                               & Not exist & \multicolumn{1}{r}{31.45} & 67435.76  & 67467.21\\ 
\hline
\multicolumn{2}{c}{Total}& \multicolumn{1}{r}{140.94} & \multicolumn{1}{r}{67443.06} & 67584.00 \\

\specialrule{.1em}{.05em}{.05em}
\end{tabular}
\end{table}
% ==============================
% End: Dataset Confusion Matrix
% ==============================

% ==================
% Evaluation Metrics
% ==================
\subsection{Evaluation Metrics}
For performance evaluation, we compute five metrics:

\subsubsection{Mean Squared Error (MSE)}
The MSE between $\hat{y}_{t}$ and $y_{t}$ is
\begin{center}
$\textrm{MSE}= \E\left [\left \| y_{t} - \hat {y}_{t} \right \|_2^2 \right ]$.
\end{center}
Our MSE criterion measures: 1) correct prediction of unit types present at each location 2) correct prediction of how many (if present).

\subsubsection{Accuracy, Precision, Recall and F1 score}
Accuracy indicates how well the existence of units is predicted. Recall reflects how much false negative rate (type 2 error) is improved. For DefogGAN perspective, type 2 error gives a more practical indicator because the damage caused by an unexpected enemy (false negative) is greater than a nonexistent enemy (false positive). Precision represents a type 1 error as a percentage of what is expected to exist. The F1 score indicates the harmonic mean of recall and precision.

% ====================
% Evaluation Interval
% ====================
\subsection{Determining Generator Training Interval}
We have experimentally determined a reasonable amount of data needed for training the DefogGAN generator. Table~\ref{tab:exp-eval1} summarizes the generator performance measured in MSE, accuracy, and F1 score computed by varying number of frames used in training. Due to the nature of the DefogGAN prediction, the MSE criterion is most valuable. The empirical results suggest training with 10-sec worth of frames the best among our tested intervals. 

% ===========================
% Table: Evaluation Interval
% ===========================
\begin{table}[h!]
\centering
\small
\caption{The DefogGAN generator performance comparison on varying number of frames used in training.}
\label{tab:exp-eval1}
\begin{tabular}{crrrrr}
\specialrule{.1em}{.05em}{.05em}
		 \multicolumn{1}{c}{Interval(frame)}        & \multicolumn{1}{c}{MSE}       & \multicolumn{1}{c}{Acc.}                & \multicolumn{1}{c}{F1}        & \multicolumn{1}{c}{Recall}  & \multicolumn{1}{c}{Preci.} \\ 
\cline{1-6}
5s(9,165) & 0.00211 & 0.99942 &  0.854 & 0.808 & 0.906 \\
10s(13,692) & \textbf{0.00208} & 0.99944 & 0.856 & 0.807 & 0.913 \\
30s(31,442) & 0.00215 & 0.99945 &  0.860 & 0.808  & \textbf{0.918} \\
60s(56,963) & 0.00213 & \textbf{0.99946} & \textbf{0.862} & 0.814 & 0.915 \\
600s(496,830) & 0.00230 & 0.99944 & 0.859 & \textbf{0.820} & 0.901 \\

%\hline\hline
\specialrule{.1em}{.05em}{.05em}
\end{tabular}
\end{table}

% =======================================
% Figure: Comparision with other methods
% =======================================
\begin{figure*}[!h]
\centering
\includegraphics[width=1\textwidth]{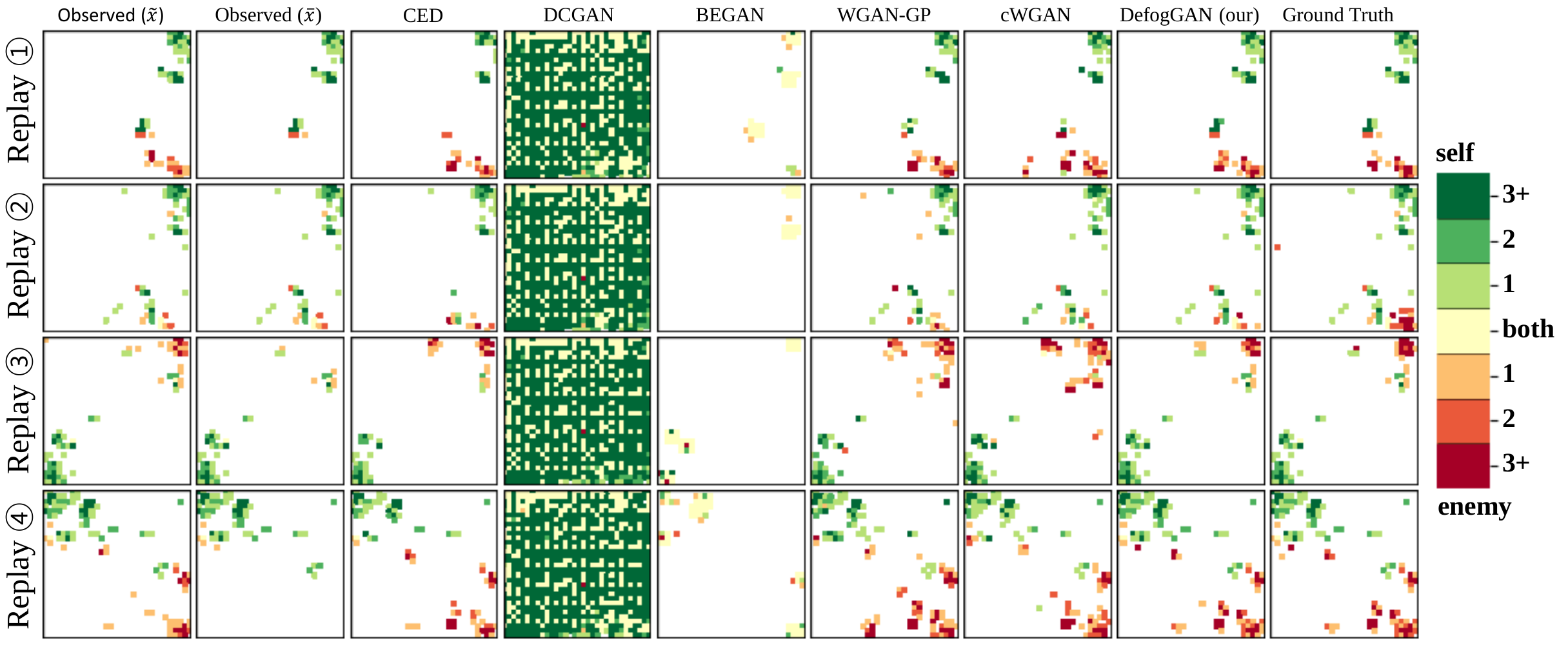}
\caption{Visualization of prediction results. On far left, we show the accumulated partially observed states ($\tilde{x}_{t}$). The second column depicts a partially observed state, $\bar{x}_{t}$. The prediction result by CED, a state-of-the-art defogger is shown in the third column. Columns 4--7 are the results by DCGAN, BEGAN, WGAN-GP and cWGAN. Our DefogGAN result is presented in the eighth column, and the ground truth on the last. Rows represent replays used for evaluation.}
\label{fig:ex3_fig} 
\end{figure*}

\subsection{Baseline}
% ================
% Table: Baseline
% ================
\begin{table}[h]
\centering
\small
\caption{The overall accuracy performance of partially observed states $\bar{x}_t$ and accumulated partially observed states $\tilde{x}_{t}$. The MSE indicates how well the units are positioned and numbered in the unit map, and the Accuracy, F1 scores, Recall and Precision indicates how well the units are aligned in the unit map.}
\label{tab:exp-eval}
\begin{tabular}{lrrrrrr}
\specialrule{.1em}{.05em}{.05em}
         & \multicolumn{1}{c}{MSE}      & \multicolumn{1}{c}{Acc.} & \multicolumn{1}{c}{F1} &\multicolumn{1}{c}{Recall} & \multicolumn{1}{c}{Preci.} \\ 
\cline{2-6}
$\bar{x}_{t}$ (partial) & 0.00548  & 0.99912 & 0.733 & 0.579 &  \textbf{1}  \\
$\tilde{x}_{t}$ (accum.) &\textbf{ 0.00370} & \textbf{0.99943} & \textbf{0.850} & \textbf{0.777}	& 0.937\\
\specialrule{.1em}{.05em}{.05em}
\end{tabular}
\end{table}

As presented in Table \ref{tab:exp-eval}, accumulated partial observation $\tilde{x}_{t}$ results in better performance than partial observation $\bar{x}_{t}$. In fact, many rule-based agents leverage $\tilde{x}_{t}$~\cite{ontanon,synnaeve1}. We take accumulated partial observation as our baseline.

% ====================
% Accuracy Comparison
% ====================
\subsection{Accuracy Comparison}
In this section, we present a comparative performance analysis for DefogGAN. A rule-based StarCraft agent using accumulated partial observation is a reasonable baseline. This baseline means that a prediction model needs to make at least better prediction than just memorizing partial observation history. For comprehensive comparison, we select a diverse range of models including an autoencoder-based model CED~\cite{synnaeve1,kahng}, simple GAN-based models, DCGAN~\cite{dcgan} and BEGAN~\shortcite{began}, and WGAN-based models, WGAN-GP~\shortcite{wgangp} and cWGAN~\cite{cwgan}.

\subsubsection{Comparison with baseline} As shown in Table \ref{tab:exp-eval3}, DefogGAN results in a 44\% decrease in MSE compared to the baseline. DefogGAN predicts the number of units in a given cell more accurately than the baseline. This is because DefogGAN is able to predict enemy units hidden in fog. On the other hand, DefogGAN seems to provide similar prediction performance in terms of accuracy and F1 score. Note that accuracy and F1 score do not measure how accurately the number of units are predicted, but just measure how accurately the existence is predicted. Then, the result can be understood that DefogGAN can predict the number of units much precisely while correctly predicting the overall distribution of units on a map.

\subsubsection{Comparison with autoencoder model} Compared to CED, one of autoencoder-based models, DefogGAN provide about 33\% decreased MSE, and about 17\% point increased F1 score. Note that recall of DefogGAN is very high, compared to that of CED. This high recall means that DefogGAN successfully discover enemy units hidden in fog. This high recall property is very important in StarCraft, since misdetected enemy units (i.e., low recall) can increase possible threat such as sudden attacks.

\subsubsection{Comparison with GAN-based models} DefogGAN makes a better prediction compared to other GAN-based models. As shown in Table \ref{tab:exp-eval3}, unconditional base GAN models such as DCGAN and BEGAN performs very poorly. This is mainly because these models are trained without reconstruction loss. WGAN-GP makes relatively good prediction results without reconstruction loss, but does not exceed DefogGAN. We carefully think that the Wasserstein distance of WGAN-GP makes an effect of reconstruction loss in training. Therefore, we do additional comparison with cWGAN, a WGAN variants that has reconstruction loss. However, cWGAN does not provide better performance than WGAN-GP. 

% =======================================
% Table: Comparision with other methods
% =======================================
\begin{table}[h!]
\centering
\small
\caption{Accuracy comparison results. DefogGAN is compared with various other models. }
\label{tab:exp-eval3}
\begin{tabular}{lrrrrrr}
\specialrule{.1em}{.05em}{.05em}
     & \multicolumn{1}{c}{MSE}     & \multicolumn{1}{c}{Acc.} & \multicolumn{1}{c}{F1} & \multicolumn{1}{c}{Recall} & \multicolumn{1}{c}{Preci.}  \\
\cline{2-6}
Baseline               & 0.00370          & 0.99943          & 0.850          &     0.777    &     \textbf{0.937}    \\
CED      & 0.00311 & 0.99896  & 0.682    & 0.538  & 0.933 \\
 DCGAN    & 2.16007 & 0.94844  & 0.019    & 0.239  & 0.010 \\
 BEGAN    & 0.01578 & 0.99353  & 0.024    & 0.039  & 0.018 \\
WGAN-GP  & 0.00348 & 0.99885  & 0.701    & 0.648  & 0.763 \\
cWGAN & 0.00372 &  0.99878 & 0.688 & 0.644 &  0.737 \\
DefogGAN & \textbf{0.00208} & \textbf{0.99944}  & \textbf{0.856}    & \textbf{0.807} & 0.913 \\

\specialrule{.1em}{.05em}{.05em}
\end{tabular}
\end{table}

\subsubsection{Visualization of prediction results} The prediction performance of DefogGAN can be effectively explained with the visualization in Figure \ref{fig:ex3_fig}. We randomly select four replays and present the defogged fully observed states predicted by each model. For example, in replay 4, we cannot see red enemy units in the lower right corner of the partially observed state $\bar{x}_{t}$. Also, we can only see a subset of enemy units from the accumulated partially observed states $\tilde{x}_{t}$. By using both observation and accumulated observation, DefogGAN generates a fully observed state $\hat{y}_t$ that looks most similar to the ground truth. Since DCGAN and BEGAN do not use reconstruction loss, they fail to generate a fully observed state that has similar pattern to the ground truth. CED generates fairly plausible full states, but DefogGAN generates more accurate results. WGAN-GP generates plausible full states without reconstruction loss. However, it seems to have a tendency to generate false positive results (i.e., low precision). cWGAN (a WGAN-GP variant that additionally use reconstruction loss) seems to reduce such false positives, but still do not make a prediction better than DefogGAN.

% ===============
% Abliation Study
% ===============
\subsection{Ablation Study}
We evaluate the performance of the proposed method that combines accumulated partial observation $\tilde{x}_t$ and partial observation $\bar{x}_t$, joint loss and reduced resolution loss. Finally, we compare the performance of the observation preserving connection.

% ======================
% Abliation Result Table
% ======================
\begin{table}[h!]
\centering
\small
\caption{Ablation study results. Each component is excluded from DefogGAN in order: current partial observation, accumulated past partial observation, adversarial loss, reconstruction loss, pyramidal loss ($L_2$ loss was used instead), and observation preserving connection.}
\label{tab:exp-eval2}
\npdecimalsign{.}
\nprounddigits{4}
\begin{tabular}{lrrrrr} 
\specialrule{.1em}{.05em}{.05em}
                                                  & \multicolumn{1}{c}{MSE.}               & \multicolumn{1}{c}{Acc}               &   \multicolumn{1}{c}{F1}        &          \multicolumn{1}{c}{ Recall}        &          \multicolumn{1}{c}{ Preci.}  \\ 
\cline{2-6}
$\circleddash X_{par}$   & 0.00293          & 0.99930          & 0.831   & \textbf{0.826}   &     0.836   \\
$\circleddash X_{acc}$    & 0.00426          & 0.99897          & 0.732    &  0.674	  &     0.802  \\
$\circleddash L_{adv}$   & 0.00310          & 0.99887          & 0.662     & 0.529	 &     0.882 \\
$\circleddash L_{rec}$     & 1.73986          & 0.97942          & 0.026     & 0.133	 &     0.015   \\
$\circleddash$ Pyramidal & 0.00210          & 0.99943          & 0.855  &     0.809  &     0.907	   \\
$\circleddash$ Ob-conn & 0.00401          & 0.99829          & 0.516  &  0.437	 &     0.631     \\
DefogGAN    & \textbf{0.00208} & \textbf{0.99944} & \textbf{0.856} & 0.807	  &     \textbf{0.913}\\
\specialrule{.1em}{.05em}{.05em}
\end{tabular}
\end{table}

DefogGAN proposed in Table \ref{tab:exp-eval2} shows that our proposed techniques in the ablation study produce good performance.

\subsubsection{Effect of concatenated partial observation} Using the concatenated partial observation method, the MSE is 29\% better than using only the accumulated partial observed information and 51\% better than using only the partial observed information. This indicates that it is important to utilize past information. In addition, when used in combination with partially observed and accumulated partially observed information, the total number of units observed from the past is identified, and certain information without type 1 errors is used for learning. In other words, it contributes to the performance improvement by showing the number of units as much as possible and the units that can be confirmed as correct.

\subsubsection{Effect of adversarial learning} The third row of Table 5 shows the overall accuracy performance of DefogGAN when trained without adversarial loss. Without adversarial loss, the overall accuracy performance significantly decreases. MSE increases about 49\% (i.e., from 0.00208 to 0.00310). F1 score decreases by 0.194 (i.e., from 0.856 to 0.662). In the area of image generation, learning with adversarial loss generates clearer images than learning with MSE loss~\cite{context,isola2017image}. In DefogGAN, we see a similar effect. We conjecture that adversarial loss also helps accurately predict the fully observed states of a game.

\subsubsection{Effect of reconstruction loss} Pyramidal reconstruction loss helps to learn fully observed states. Since it measures the difference between a predicted fully observed state and the ground truth at multiple scales, it helps DefogGAN accurately predict the total number of units hidden in the fog.

\subsubsection{Effect of observation preserving connection} As shown in the 6th row of Table \ref{tab:exp-eval2}, when trained without observation preserving connection, the overall accuracy performance of DefogGAN significantly decreases. More specifically, MSE increases about 200\% (i.e., from 0.00208 to 0.00410). F1 score decreases by 0.340 (i.e., from 0.856 to 0.516). This can be considered as a similar effect that skip connection of U-Net~\cite{unet} provides better results by allowing information to flow from input to output.

\section{Conclusion}
We have presented DefogGAN, a generative approach for game AI to predict crucial state information unavailable due to the fog of war. DefogGAN accurately generates defogged images of a game that can be used to improve win rates against expert human players. In our experiments with StarCraft, we have validated that DefogGAN achieves a superior performance against state-of-the-art defoggers. Improving on imperfect information during an RTS game play could bring substantially better macro-management overall, although this is an ongoing research area for game AI. DefogGAN is one of the earliest applications for adversarial learning to improve the fog of war problem, and it can be applied to other real-world POMDP problems.

% acknowledgment
\section{Acknowledgment}
The authors would like to thank Dr. Wonpyo Hong, CEO and President of Samsung SDS, whose vision in AI has led to create SAIDA Lab for advanced AI research.

% references 
\bibliographystyle{aaai}
\bibliography{paper}
\end{document}